\documentclass{article}
\usepackage[preprint]{colm2026_conference}

\usepackage{microtype}
\usepackage{url}
\usepackage{booktabs}
\usepackage{amsmath}
\usepackage{amssymb}
\usepackage{float}
\usepackage{multirow}
\usepackage{graphicx}
\usepackage{subcaption}
\usepackage{xcolor}

\usepackage{hyperref}
\definecolor{darkblue}{rgb}{0, 0, 0.5}
\hypersetup{colorlinks=true, citecolor=darkblue, linkcolor=darkblue, urlcolor=darkblue}

\title{Generalizable Probes for Detecting Factual Correctness in LLMs}

\author{Smitali Bhandari\thanks{Primary contributors.}\kern0.45em, Harshith M R\footnotemark[1]\kern0.45em, Jayden Koshy Joe\thanks{Correspondence to \texttt{jaydenkjoe@gmail.com}.}\kern0.45em,\\
\textbf{Sharan Keshav, Arsh Mathur} \\
Indian Institute of Technology Madras
}

\begin{document}

\maketitle

\begin{abstract}
Language models are often reported to hallucinate factual information. Activation probes offer a lightweight way to detect them, yet probes trained on one benchmark often fail to generalize to others. In this work, we introduce a balanced four-class labelling scheme which improves cross-dataset performance. Across three instruction-tuned LLMs and three QA datasets with up to 6.2K test examples, non-linear probes trained solely on TriviaQA generalize zero-shot to HaluEval and PopQA, outperforming conventional binary end-of-sequence probes in five of six out-of-distribution settings. We also show that non-linear probes outperform linear baselines. We hope that our findings will enable more generally usable probes in detecting factual recall errors.
\end{abstract}

\section{Introduction}\label{sec:intro}
Large Language Models (LLMs) frequently generate fluent but factually incorrect answers, known as hallucinations~\citep{ji2023survey, zhang2023siren, huang2023survey}. A prominent line of interpretability research detects these errors by training probes on internal activations~\citep{azaria2023internal, burns2023discovering, marks2023geometry, orgad2024llms}. Reliable activation probes would need to generalize beyond their training benchmark; however, it has been reported that probes trained on one dataset often degrade significantly in out-of-distribution settings~\citep{levinstein2023still, orgad2024llms, dubanowska2025representation, cho2026confidence}. This motivates our central question: \textbf{how can we train probes that reliably detect hallucinations such that they generalize across datasets?}

Three hypotheses could explain why existing probes struggle to transfer. First, \emph{label confounders}: under standard binary correct/incorrect labels on a model's own free-form answers, correctness correlates with surface cues such as answer length or phrasing~\citep{orgad2024llms} and benchmark artifacts~\citep{hussain2026parallax}. Probes can exploit these dataset-specific shortcuts in distribution and fail under shift. Second, \emph{insufficient probe complexity}: following the Linear Representation Hypothesis~\citep{park2023linear, marks2023geometry}, most detection and steering methods assume correctness lies along a single linear direction~\citep{zou2023representation, li2023inference, lee2026meanshift}, which may miss non-linear or multi-dimensional structure. Third, \emph{lack of a transferable representation}: LLMs may simply not encode a domain-general signal of whether their own answers are correct~\citep{orgad2024llms, poulis2026testing}.

Our experiments test all three hypotheses across three instruction-tuned models (Gemma-2-2b-it, Llama-3.2-3B-Instruct, and Qwen2.5-3B-Instruct) and three QA datasets (TriviaQA, HaluEval, and PopQA) with 20K queries in train and up to \textbf{6.2K test examples}. We find that LLMs do encode a transferable internal correctness signal, ruling out the third hypothesis, and that addressing both label confounders and probe complexity is necessary to read it reliably. Specifically, we make two core contributions:
\begin{itemize}
    \item \textbf{Controlling answer-level confounders unlocks cross-dataset generalization} (Sections~\ref{sec:method}, \ref{ssec:cross-dataset}, and~\ref{ssec:confabulation}): Addressing the confounder and representation hypotheses, we introduce a balanced four-class labelling scheme over model-generated answers that eliminates answer-only shortcuts by construction. These probes transfer significantly ($p \leq 0.001$) to HaluEval (0.68--0.74 AUROC) and PopQA (0.57--0.68 AUROC) across all three models, beating binary EOS probes in five of six OOD settings.
    \item \textbf{A single linear direction misses transferable correctness signal} (Sections~\ref{ssec:main_results}, \ref{ssec:cross-dataset}, and~\ref{sec:subspace}): Addressing probe complexity, small non-linear probes beat full-data and shrinkage LDA by 0.05--0.10 AUROC in distribution and lead in five of six OOD settings. SVD analysis on Gemma-2-2b-it further shows that while low-dimensional projections fit TriviaQA well, long-tail transfer to PopQA drops sharply below 256 directions and vanishes by 16.
\end{itemize}

\section{Related Work}\label{sec:related_work}
\paragraph{Truthfulness probing and label construction.}
A rich body of work probes language models for internal signals of truthfulness. \citet{kadavath2022language} append ``True''/``False'' prompts to model answers and read the output probability $P(\text{True})$, while \citet{azaria2023internal} train binary MLP probes (SAPLMA) on hidden states of curated statements. \citet{burns2023discovering} introduces Contrast-Consistent Search (CCS), which appends ``True''/``False'' contrast tokens and searches unsupervised for a linear direction satisfying negation consistency. Our four-class scheme builds on this evaluation-token idea in a distinct setting. Unlike $P(\text{True})$, we probe internal residual-stream activations rather than output probabilities. Unlike SAPLMA and CCS, we probe the model's own free-form QA answers and supervise probes on balanced four-class agreement labels so answer-only features carry no information about the target. Recent probes on model-generated answers~\citep{orgad2024llms, kossen2024semantic, obeso2025realtime} use binary correctness labels. Prior studies also note that binary probing setups can be confounded by answer length~\citep{orgad2024llms}, query-level cues~\citep{slobodkin2023curious}, spurious dataset correlations~\citep{dubanowska2025representation}, and answer-bearing context passages~\citep{hussain2026parallax}. Our labelling scheme directly addresses these answer-level confounders.

\paragraph{Cross-dataset generalization.}
Transferring truthfulness probes across datasets remains challenging. \citet{levinstein2023still} observe that SAPLMA and CCS struggle under basic shifts, and \citet{orgad2024llms} report limited cross-dataset transfer for single-dataset error detectors. Later evaluations similarly find weak out-of-distribution transfer~\citep{dubanowska2025representation}, sensitivity to paraphrases and typos~\citep{haller2025brittle}, and shifts in truth geometry across tasks and prompts~\citep{poulis2026testing}. Several promising approaches target this gap. \citet{liu2024universal} and \citet{cho2026confidence} show that joint training across diverse datasets improves cross-dataset transfer, \citet{kossen2024semantic} train probes to predict semantic entropy as a generalizable uncertainty proxy, and \citet{obeso2025realtime} show cross-model transfer on long-form entities at larger scales. Complementary to multi-dataset training and scale, we test whether removing answer-level confounders allows a probe trained on a single QA dataset to transfer zero-shot to unseen benchmarks.

\paragraph{Linearity, covariance estimation, and subspace geometry.}
Whether factual correctness is encoded along a single linear direction~\citep{mikolov2013linguistic, park2023linear, marks2023geometry, zou2023representation, li2023inference} or spans a richer non-linear and multi-dimensional space~\citep{elhage2022superposition, engels2024not} is being actively studied. \citet{aiersilan2026linearly} find hallucinations linearly decodable from mid-layer states, and \citet{lee2026meanshift} argue that the hallucination signal is a single mean-shift direction where L2 logistic regression matches complex probes and shrinkage LDA closes most of the linear--non-linear gap. Conversely, \citet{liang2025neural} report gains from lightweight MLPs over linear probes, while \citet{burger2024truth}, \citet{cho2026confidence}, \citet{vicidomini2026anatomy}, and \citet{bao2025probing} find multi-dimensional truth subspaces whose geometry varies with knowledge and scale. We add controlled evidence by comparing MLPs against full-data LDA (both with, and without shrinkage) under confounder-controlled labels and measure how subspace dimensionality affects out-of-distribution transfer.

\section{Methodology}\label{sec:method}

\paragraph{Overview.}
We train probes on residual-stream activations to detect whether an LLM's own generated answers are factually correct. Section~\ref{ssec:representation} introduces our four-class labelling scheme for controlling answer-level confounders. Section~\ref{ssec:dataset} describes the models, datasets, generation procedure, and splits used to evaluate in-distribution and cross-dataset generalization. Section~\ref{ssec:probing} details our probes, baselines, validation-only layer selection, permutation tests, and SVD projections.

\subsection{Confounder-Controlled Four-Class Representation}\label{ssec:representation}

For a probe to reflect a model's internal knowledge of truthfulness, it must not succeed via surface confounders such as answer length~\citep{orgad2024llms} that merely correlate with correctness. Let $\mathcal{M}$ be an LLM, $q$ a query, $y$ the reference answer, and $\hat{y} \sim \mathcal{M}(q)$ a sampled generation. We define correctness $c \in \{0,1\}$, where $c=1$ if $\hat{y}$ matches $y$ under fuzzy string matching and $c=0$ otherwise (label quality is audited in Appendix~\ref{app:grounding}).

We append an evaluation token $t \in \{\text{``True''}, \text{``False''}\}$ to form $x_t = [q;\, \hat{y};\, t]$. The target label $z \in \{0, 1\}$ indicates whether $t$ agrees with the answer's actual correctness:
\[
z = \begin{cases} c & \text{if } t = \text{``True''}, \\ 1 - c & \text{if } t = \text{``False''}. \end{cases}
\]
This yields four input sequence classes (Table~\ref{tab:input_sequence_classes}): CT and WF carry $z=1$ (agreement), while CF and WT carry $z=0$ (disagreement).

\begin{table}[h]
\centering
\footnotesize
\renewcommand{\arraystretch}{0.90}
\begin{tabular}{llccc}
\toprule
\textbf{Class} & \textbf{Description} & $c$ & $t$ & $z$ \\
\midrule
CT & Correct generation labeled as ``True'' & 1 & ``True'' & 1 \\
CF & Correct generation labeled as ``False'' & 1 & ``False'' & 0 \\
WT & Wrong generation labeled as ``True'' & 0 & ``True'' & 0 \\
WF & Wrong generation labeled as ``False'' & 0 & ``False'' & 1 \\
\bottomrule
\end{tabular}
\caption{The four balanced input sequence classes, ensuring $P(z{=}1 \mid q, \hat{y}) = 0.5$ and $P(z{=}1 \mid t) = 0.5$.}
\label{tab:input_sequence_classes}
\end{table}

Balancing CT, CF, WT, and WF removes two shortcut families by construction. Any feature of the query and answer alone---such as answer length, style, or query priors---predicts $z$ only at chance, since both correct and wrong generations are split evenly across $z=1$ and $z=0$. The appended token $t$ alone is likewise uninformative. To exceed chance, a probe must read the interaction between $t$ and the model's internal representation of whether $\hat{y}$ is correct.

\subsection{Dataset Construction}\label{ssec:dataset}

We evaluate three open-weight instruction-tuned LLMs: Gemma-2-2b-it~\citep{team2024gemma}, Llama-3.2-3B-Instruct~\citep{dubey2024llama} (Llama-3.2-3b), and Qwen2.5-3B-Instruct~\citep{qwen2025qwen25technicalreport} (Qwen-2.5-3b). We use three QA benchmarks: TriviaQA~\citep{joshi2017triviaqa}, HaluEval~\citep{li2023halueval}, and PopQA~\citep{mallen2023popqa}. All probes are trained solely on TriviaQA, while HaluEval and PopQA serve as unseen zero-shot test sets. For HaluEval, we provide only the question and exclude the accompanying knowledge passage.

\paragraph{Generation and activation extraction.}
For each query $q_i$, we sample 32 independent responses using nucleus sampling ($T{=}0.7$, $p{=}0.9$) and retain unique responses. Sampling multiple responses captures natural variance in model outputs and yields both correct and wrong answers for many queries. For each retained generation $\hat{y}_j$, we extract residual-stream activations $\phi_\ell \in \mathbb{R}^d$ at every layer $\ell \geq 10$ at two positions: the \emph{T/F position}, taken at the appended ``True'' or ``False'' token of $x_t = [q;\, \hat{y};\, t]$ (our primary four-class setting), and the \emph{EOS position}, taken at the final token of $[q;\, \hat{y}]$ without an evaluation token (used for binary baselines and cross-position transfer).

\paragraph{Balancing and splits.}
We subsample training and validation sets so that CT, CF, WT, and WF are equally represented, and balance EOS sets between correct and incorrect generations. TriviaQA training, validation, and test sets come from disjoint splits of 20K, 2K, and 2K queries. This gives 643K T/F training pairs per layer for Gemma-2-2b-it and 1.22M each for Llama-3.2-3b and Qwen-2.5-3b, with 120K--127K validation pairs per model. In every test set (TriviaQA, HaluEval, PopQA), we keep at most one correct and one wrong generation per query so no query dominates evaluation, yielding 1.2K--6.2K test examples per dataset. Full counts and a label-quality audit of clarification requests appear in Appendix~\ref{app:data}.

\subsection{Probing Methods and Evaluation Protocol}\label{ssec:probing}

\paragraph{Non-linear and linear probes.}
Our non-linear probe is a three-hidden-layer MLP ($d \to 256 \to 64 \to 32 \to 1$) with LeakyReLU activations, trained independently at each layer using binary cross-entropy, AdamW, cosine decay, Optuna hyperparameter tuning, and validation early stopping (averaged over three seeds for Gemma-2-2b-it T/F; Appendix~\ref{app:training}). We compare against three linear baselines on identical balanced splits: (1) L2-regularized logistic regression on standardized activations, (2) Difference-of-Means (DoM), $w_{\text{DoM}} = \bar{\phi}_{z=1} - \bar{\phi}_{z=0}$, on raw activations, and (3) Linear Discriminant Analysis (LDA), $w = \hat{\Sigma}^{-1}(\bar{\phi}_{z=1} - \bar{\phi}_{z=0})$, fitted on the full training set ($n/d \in [280, 600]$). Following \citet{lee2026meanshift}, we also test Ledoit--Wolf~\citep{ledoit2004well} and validation-tuned covariance shrinkage (Appendix~\ref{app:lda_shrinkage}).

\paragraph{Layer selection, significance, and subspace probing.}
For each model and probe, we select a single layer maximizing TriviaQA validation AUROC and reuse that exact layer across all test sets. Each test AUROC is paired with a one-sided 1,000-shuffle permutation test, $p = (1 + \#\{\text{shuffled AUROC} \geq \text{observed}\})/1001$, and results with $p > 0.001$ are marked not significant (n.s.). To measure subspace dimensionality (Section~\ref{sec:subspace}), we project activations onto the top-$k$ right singular vectors of each layer's training matrix ($k \in \{576, 256, \dots, 2\}$) and retrain the MLP (Appendix~\ref{app:svd_details}).

\section{Results}\label{sec:results}

\paragraph{Overview.}
Section~\ref{ssec:main_results} evaluates in-distribution detection on TriviaQA across layers, testing whether a confounder-free signal exists and how non-linear probes compare against strong linear baselines. Section~\ref{ssec:cross-dataset} tests zero-shot transfer to HaluEval and PopQA. Section~\ref{ssec:confabulation} directly compares our four-class T/F probes against conventional binary EOS probes to isolate the role of confounder control in out-of-distribution generalization. Section~\ref{ssec:eos_transfer} tests whether T/F-trained probes transfer to EOS activations when no evaluation token is present.

\subsection{In-Distribution Detection and the Linear--Non-Linear Gap}\label{ssec:main_results}
\begin{figure}[H]
    \centering
    \includegraphics[width=1\textwidth]{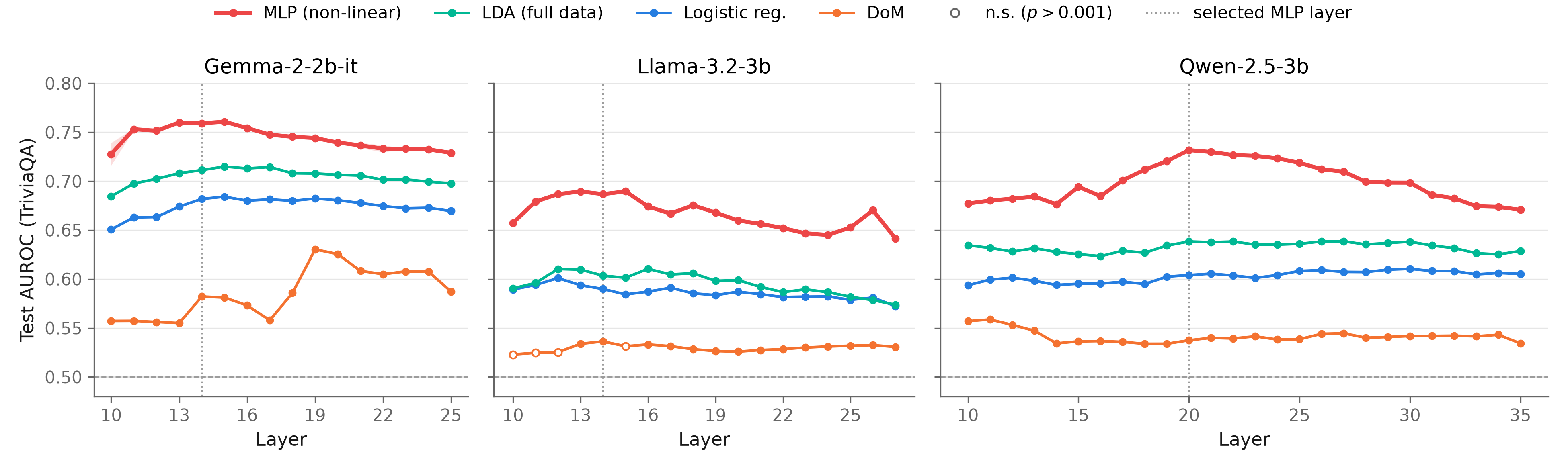}
    \caption{Layer-wise test AUROC on TriviaQA T/F activations for Gemma-2-2b-it (MLP band: $\pm 1$ SD over 3 seeds), Llama-3.2-3b, and Qwen-2.5-3b. Hollow markers indicate $p > 0.001$ (n.s.) and dotted vertical lines mark the selected MLP layer. Across all three models, the MLP outperforms every linear baseline at every layer $\geq 10$ and peaks near mid-depth.}
    \label{fig:layerwise_accuracy}
\end{figure}

Figure~\ref{fig:layerwise_accuracy} shows layer-wise test AUROC on TriviaQA T/F activations across all three models, and Table~\ref{tab:main_results} compares the MLP against the strongest linear baseline (LDA in all three models).

\begin{table}[H]
\centering
\footnotesize
\renewcommand{\arraystretch}{0.92}
\begin{tabular}{ll c c ccc}
\toprule
\multirow{2}{*}{\textbf{Model}} & \multirow{2}{*}{\textbf{Probe}} & \multirow{2}{*}{\textbf{Layer}}
  & \textbf{Val} & \multicolumn{3}{c}{\textbf{Test (TriviaQA)}} \\
\cmidrule(lr){4-4} \cmidrule(lr){5-7}
 & & & AUROC & Acc & AUROC & F1 \\
\midrule
\multirow{2}{*}{\texttt{Gemma-2-2b-it}}
 & Non-linear (MLP) & 14 & \textbf{0.8135} & \textbf{0.6930} & \textbf{0.7591} & \textbf{0.6876} \\
 & Linear (LDA)     & 14 & 0.7672 & 0.6494 & 0.7113 & 0.6491 \\
\midrule
\multirow{2}{*}{\texttt{Llama-3.2-3b}}
 & Non-linear (MLP) & 14 & \textbf{0.7628} & \textbf{0.6333} & \textbf{0.6867} & \textbf{0.6159} \\
 & Linear (LDA)     & 16 & 0.6557 & 0.5787 & 0.6104 & 0.6011 \\
\midrule
\multirow{2}{*}{\texttt{Qwen-2.5-3b}}
 & Non-linear (MLP) & 20 & \textbf{0.7916} & \textbf{0.6664} & \textbf{0.7316} & \textbf{0.6747} \\
 & Linear (LDA) & 11 & 0.6959 & 0.5967 & 0.6318 & 0.5956 \\
\bottomrule
\end{tabular}
\caption{In-distribution TriviaQA T/F results. The linear row shows the strongest linear baseline (LDA in all three models). All test AUROCs satisfy $p \leq 0.001$; Gemma-2-2b-it MLP values are averaged over three seeds.}
\label{tab:main_results}
\end{table}

First, LLMs encode a confounder-free correctness signal. Across all three models, MLP probes achieve significant above-chance performance on TriviaQA ($p \leq 0.001$; Table~\ref{tab:main_results}).

Second, non-linear probes consistently outperform all linear baselines across every model and layer (Figure~\ref{fig:layerwise_accuracy}, Table~\ref{tab:main_results}). Prior work suggests that apparent non-linear gains over linear probes can arise when a linear classifier's covariance matrix is poorly estimated~\citep{lee2026meanshift}. Because our LDA baseline uses the entire training set, covariance estimation error is small, and adding Ledoit--Wolf or tuned covariance shrinkage leaves LDA performance virtually unchanged (Appendix~\ref{app:lda_shrinkage}). A single linear direction therefore captures much, but not all, of the available correctness signal.

Third, probe performance peaks near the middle of the network (roughly half of model depth) across all three architectures, consistent with prior accounts of mid-layer factual recall~\citep{geva2023dissecting}.

\subsection{Cross-Dataset Generalization}\label{ssec:cross-dataset}

We next evaluate TriviaQA-trained probes zero-shot on HaluEval and PopQA without retraining (Table~\ref{tab:cross_dataset}; layer-wise transfer curves appear in Appendix~\ref{app:additional_layerwise}).

\begin{table}[H]
\centering
\footnotesize
\renewcommand{\arraystretch}{0.90}
\begin{tabular}{ll c c cc}
\toprule
\multirow{2}{*}{\textbf{Model}} & \multirow{2}{*}{\textbf{Probe}} & \multirow{2}{*}{\textbf{Layer}}
  & \textbf{Val AUROC} & \multicolumn{2}{c}{\textbf{Test AUROC (Zero-Shot)}} \\
\cmidrule(lr){4-4} \cmidrule(lr){5-6}
 & & & TriviaQA & HaluEval & PopQA \\
\midrule
\multirow{4}{*}{\texttt{Gemma-2-2b-it}}
 & Non-linear (MLP) & 14 & 0.8135 & \textbf{0.7447} & \textbf{0.6525} \\
 & Linear (LogReg)  & 15 & 0.7276 & 0.6666 & n.s. \\
 & Linear (LDA)     & 14 & 0.7672 & 0.7032 & 0.5576 \\
 & Linear (DoM)     & 20 & 0.6569 & 0.6158 & n.s. \\
\midrule
\multirow{4}{*}{\texttt{Llama-3.2-3b}}
 & Non-linear (MLP) & 14 & 0.7628 & \textbf{0.7095} & \textbf{0.6778} \\
 & Linear (LogReg)  & 20 & 0.6248 & 0.6332 & 0.6485 \\
 & Linear (LDA)     & 16 & 0.6557 & 0.6304 & 0.6614 \\
 & Linear (DoM)     & 26 & 0.5517 & 0.5972 & 0.5893 \\
\midrule
\multirow{4}{*}{\texttt{Qwen-2.5-3b}}
 & Non-linear (MLP) & 20 & 0.7916 & \textbf{0.6779} & 0.5731 \\
 & Linear (LogReg)  & 11 & 0.6520 & 0.6116 & 0.5913 \\
 & Linear (LDA)     & 11 & 0.6959 & 0.6431 & 0.5954 \\
 & Linear (DoM)     & 11 & 0.5406 & 0.6705 & \textbf{0.6716} \\
\bottomrule
\end{tabular}
\caption{Zero-shot cross-dataset transfer: TriviaQA-trained T/F probes tested without retraining on HaluEval and PopQA. All numeric test entries satisfy $p \leq 0.001$.}
\label{tab:cross_dataset}
\end{table}

Our four-class MLPs transfer significantly ($p \leq 0.001$) in all six out-of-distribution settings (Table~\ref{tab:cross_dataset}). Across probe families, the MLP achieves the highest AUROC in five of six settings, consistently outperforming all the linear baselines. On Gemma-2-2b-it PopQA, both logistic regression and DoM fail to reach significance.

The sole exception is Qwen-2.5-3b on PopQA, where all three linear probes exceed the MLP, most clearly DoM, despite DoM performing weakly on TriviaQA validation. PopQA is also affected by a label-distribution shift: PopQA asks about ambiguous long-tail entities, and models ask for clarification instead of answering far more often than in TriviaQA (Appendix~\ref{app:grounding}). Fuzzy matching marks these clarifications as incorrect.

\subsection{Four-Class vs.\ Conventional Binary Probes}\label{ssec:confabulation}

To test whether four-class confounder control is responsible for this cross-dataset transfer, Table~\ref{tab:confabulation} compares our four-class T/F MLPs against conventional binary MLPs trained on EOS activations to predict correctness $c$ directly (layer-wise plots are in Appendix~\ref{app:additional_layerwise}).

\begin{table}[H]
\centering
\footnotesize
\renewcommand{\arraystretch}{0.90}
\begin{tabular}{ll cc cc}
\toprule
\multirow{2}{*}{\textbf{Model}} & \multirow{2}{*}{\textbf{Dataset}}
  & \multicolumn{2}{c}{\textbf{T/F (4-class)}} & \multicolumn{2}{c}{\textbf{EOS (binary)}} \\
\cmidrule(lr){3-4} \cmidrule(lr){5-6}
 & & Val AUROC & Test AUROC & Val AUROC & Test AUROC \\
\midrule
\multirow{3}{*}{\texttt{Gemma-2-2b-it}}
 & TriviaQA (ID)  & \multirow{3}{*}{0.8135 (L14)} & 0.7591 & \multirow{3}{*}{0.8214 (L13)} & \textbf{0.7689} \\
 & HaluEval (OOD) &                              & \textbf{0.7447} &                              & 0.6536 \\
 & PopQA (OOD)    &                              & \textbf{0.6525} &                              & 0.5413 \\
\midrule
\multirow{3}{*}{\texttt{Llama-3.2-3b}}
 & TriviaQA (ID)  & \multirow{3}{*}{0.7628 (L14)} & 0.6867 & \multirow{3}{*}{0.8294 (L12)} & \textbf{0.7519} \\
 & HaluEval (OOD) &                              & \textbf{0.7095} &                              & 0.6940 \\
 & PopQA (OOD)    &                              & \textbf{0.6778} &                              & 0.6415 \\
\midrule
\multirow{3}{*}{\texttt{Qwen-2.5-3b}}
 & TriviaQA (ID)  & \multirow{3}{*}{0.7916 (L20)} & \textbf{0.7316} & \multirow{3}{*}{0.6783 (L10)} & 0.6154 \\
 & HaluEval (OOD) &                              & \textbf{0.6779} &                              & 0.6066 \\
 & PopQA (OOD)    &                              & 0.5731 &                              & \textbf{0.5853} \\
\bottomrule
\end{tabular}
\caption{Four-class T/F vs.\ conventional binary EOS MLP probes trained on TriviaQA. All entries satisfy $p \leq 0.001$. Binary probes match or beat four-class probes in distribution (ID) on Gemma and Llama, but four-class probes win in five of six OOD settings.}
\label{tab:confabulation}
\end{table}

The contrast between in-distribution and out-of-distribution performance directly supports the confounder hypothesis (Table~\ref{tab:confabulation}). In distribution on TriviaQA, binary EOS probes beat four-class T/F probes on Gemma-2-2b-it and Llama-3.2-3b. Out of distribution, however, this pattern reverses: four-class probes outperform binary EOS probes in five of six settings---on HaluEval for every model, and on PopQA for Gemma-2-2b-it and Llama-3.2-3b---with only Qwen-2.5-3b PopQA slightly favoring EOS. This reversal is consistent with binary labels letting probes exploit dataset-specific answer cues that inflate in-distribution accuracy but fail under shift. Because the two setups also differ in token position (T/F vs.\ EOS), isolating the pure label effect at a fixed token remains an important follow-up (Section~\ref{sec:discussion}).

\subsection{Transfer Across Token Positions}\label{ssec:eos_transfer}

We also evaluate TriviaQA T/F-trained probes without retraining on TriviaQA EOS activations (where no evaluation token is appended). On Gemma-2-2b-it, the signal transfers cleanly across positions: the MLP retains strong above-chance performance ($p \leq 0.001$), well ahead of all linear baselines. However, this cross-position alignment does not hold across all models. Transfer to EOS is near chance on Llama-3.2-3b and not significant for any probe on Qwen-2.5-3b (full table and layer-wise curves in Appendix~\ref{app:eos_transfer_other}).

\section{How Many Dimensions Does the Transferable Signal Need?}\label{sec:subspace}

Having shown that a single linear direction misses transferable correctness signal, we ask how many dimensions are needed to preserve in-distribution versus cross-dataset performance. On Gemma-2-2b-it T/F activations ($d{=}2304$), we project each layer onto the top-$k$ right singular vectors of its TriviaQA training matrix ($k \in \{576, 256, \dots, 2\}$) and retrain the MLP probe.

\begin{figure}[t]
    \centering
    \includegraphics[width=0.54\textwidth]{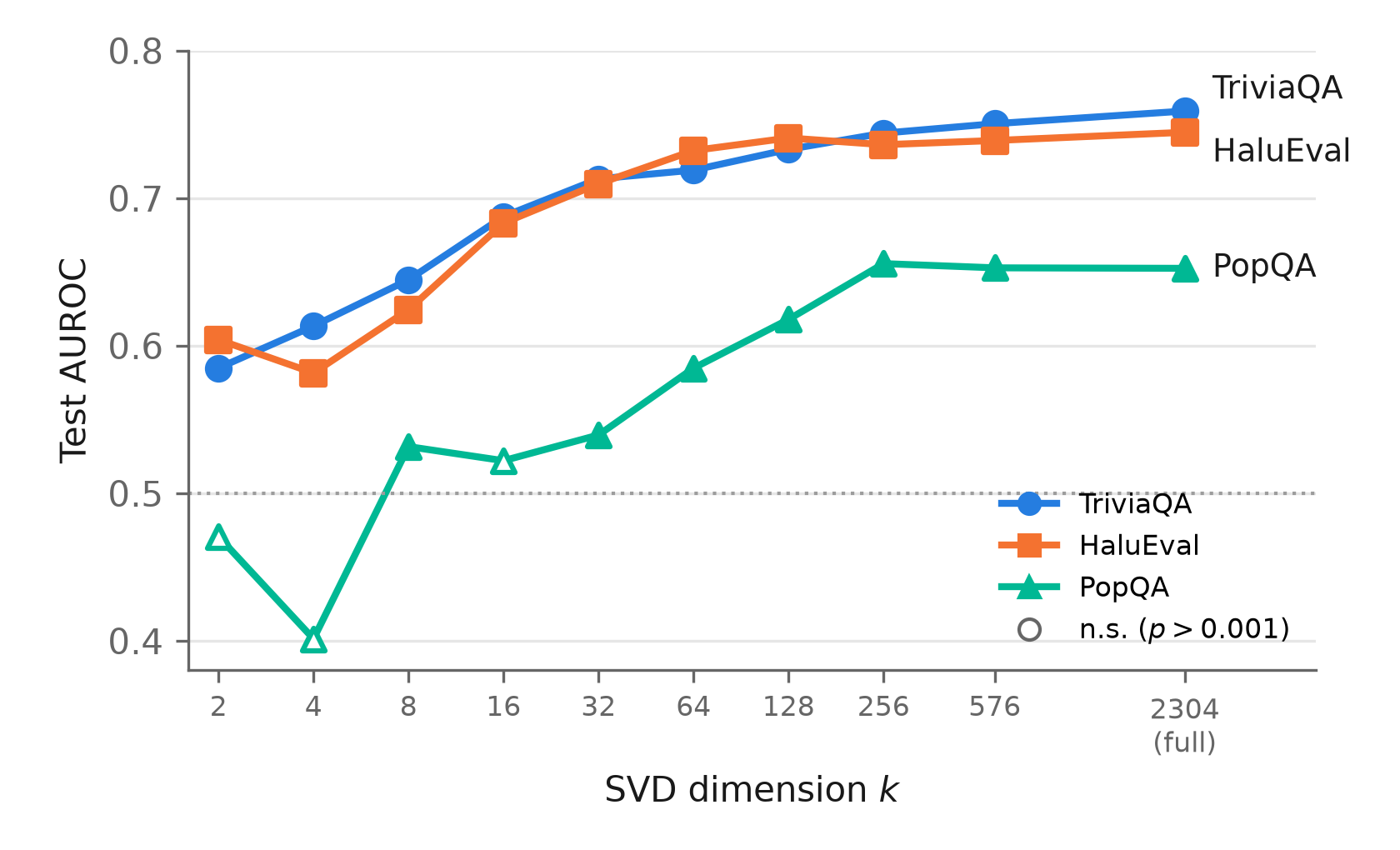}
    \caption{Test AUROC of SVD-projected MLP probes on Gemma-2-2b-it vs.\ projection dimension $k$ (trained on TriviaQA T/F, zero-shot on HaluEval and PopQA; 2304 is full space). Hollow markers: n.s.\ ($p > 0.001$). TriviaQA and HaluEval retain most signal down to $k{=}16\text{--}64$, while PopQA collapses below $k{=}256$ and vanishes by $k{=}16$.}
    \label{fig:svd_vs_k}
\end{figure}

\begin{table}[H]
\centering
\footnotesize
\renewcommand{\arraystretch}{0.83}
\begin{tabular}{r c c ccc}
\toprule
\textbf{$k$} & \textbf{Layer} & \textbf{Val (TriviaQA)} & \textbf{TriviaQA} & \textbf{HaluEval} & \textbf{PopQA} \\
\midrule
2304 (full) & 14 & 0.8135 & 0.7591 (100\%) & 0.7447 (100\%) & 0.6525 (100\%) \\
576         & 14 & 0.8075 & 0.7505 (97\%)  & 0.7391 (98\%)  & 0.6528 (100\%) \\
256         & 14 & 0.8012 & 0.7441 (94\%)  & 0.7363 (97\%)  & 0.6557 (102\%) \\
128         & 14 & 0.7943 & 0.7331 (90\%)  & 0.7408 (98\%)  & 0.6179 (77\%)  \\
64          & 14 & 0.7803 & 0.7191 (85\%)  & 0.7323 (95\%)  & 0.5848 (56\%)  \\
32          & 17 & 0.7630 & 0.7131 (82\%)  & 0.7098 (86\%)  & 0.5395 (26\%)  \\
16          & 17 & 0.7254 & 0.6877 (72\%)  & 0.6832 (75\%)  & n.s.           \\
8           & 25 & 0.6739 & 0.6447 (56\%)  & 0.6244 (51\%)  & 0.5317 (21\%)  \\
4           & 13 & 0.6371 & 0.6134 (44\%)  & 0.5816 (33\%)  & n.s.           \\
2           & 16 & 0.6002 & 0.5849 (33\%)  & 0.6041 (43\%)  & n.s.           \\
\bottomrule
\end{tabular}
\caption{SVD-projected MLP probes on Gemma-2-2b-it T/F activations at TriviaQA-validation-selected layers. Percentages show retained above-chance AUROC: $(\text{AUROC}_k - 0.5)/(\text{AUROC}_{\text{full}} - 0.5)$. Numeric test entries satisfy $p \leq 0.001$; n.s.: $p > 0.001$.}
\label{tab:svd_generalization}
\end{table}

Figure~\ref{fig:svd_vs_k} and Table~\ref{tab:svd_generalization} show a sharp gap between in-distribution retention and long-tail transfer (all layers appear in Appendix~\ref{app:svd_layerwise}). On TriviaQA and HaluEval, performance degrades smoothly as $k$ shrinks: a few dozen leading directions retain most of the full-space probe's above-chance signal, and even $k{=}2$ remains significant on TriviaQA. On PopQA, however, transfer holds down to $k{=}256$ but drops sharply below that point and disappears for $k \leq 16$, even inverting below chance at $k{=}4$. Gradient-based SHAP attribution~\citep{lundberg2017unified} on the $k{=}576$ probe (Appendix~\ref{app:shap}) agrees: inputs share a small core of dominant directions far above chance, while remaining attribution spreads across roughly half the subspace. In-distribution evaluation alone can thus overstate how much transferable signal a low-dimensional projection preserves.

\section{Discussion, Limitations, and Conclusion}\label{sec:discussion}

Across three instruction-tuned LLMs, we address the hypotheses introduced. First, we show that activations in LLMs can be used to detect correctness. Second, controlling answer-level confounders via balanced four-class labeling allows for dataset transfer: binary EOS probes match or beat four-class probes in distribution on two models, yet four-class probes win in five of six OOD settings. Third, small MLPs outperform LDA (with and without shrinkage) in distribution and lead in five of six OOD settings. Finally, we also show with SVD probing on Gemma-2-2b-it that long-tail PopQA transfer relies on substantially higher-dimensional structure than in-distribution tests suggest.

\paragraph{Limitations and future work.}
Our study has been constrained to models with $<$3B parameters. Previous work suggests that truth may become more linearly separable in larger models~\citep{marks2023geometry, bao2025probing, obeso2025realtime}. Testing our four-class protocol and LDA baselines at larger scale is a natural next step. Four-class T/F and binary EOS probes differ in both labels and token position, so training binary probes at the evaluation token would isolate the pure label effect. It also remains to be checked whether the MLP probes perform better by extracting a signal associated with correctness from the LLM, or implicitly perform an XOR-like operation, leaving them still susceptible to confounders.

\section*{Acknowledgments}
The authors thank ExceptionRaised! and GradCapital for funding support towards this work. The authors thank Dr.\ Nitish Mital for guidance, and Pakshal Nagda, Vedant Misra, Saahil Faraaz Shaikh, Sriram V, Samrudh Raaj and Eshika Nahata for valuable discussions.

\bibliography{colm2026_conference}
\bibliographystyle{colm2026_conference}

\clearpage
\appendix

\section{Additional Layer-wise Plots}\label{app:additional_layerwise}

\begin{figure}[H]
    \centering
    \includegraphics[width=\textwidth]{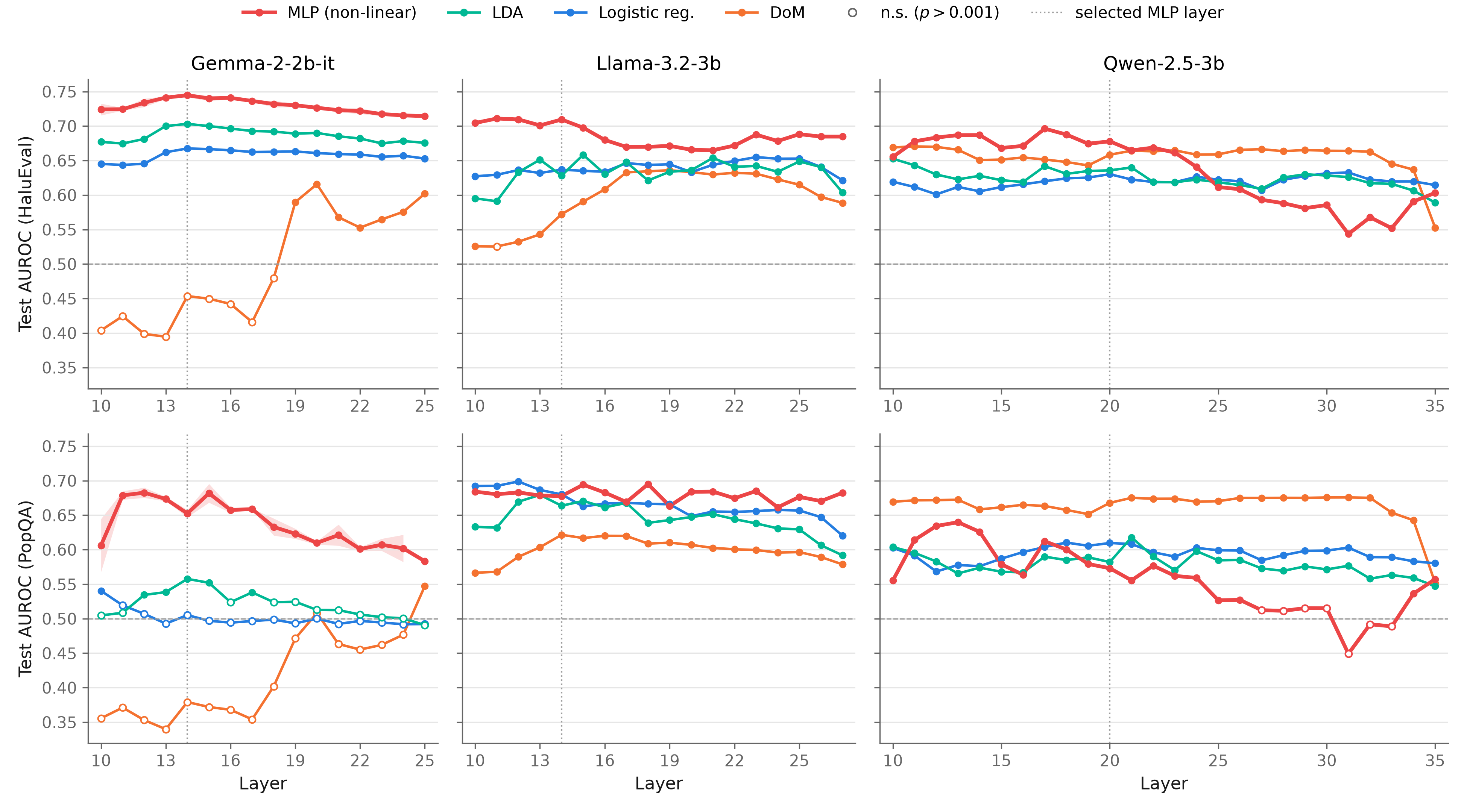}
    \caption{Layer-wise test AUROC of probes trained on TriviaQA T/F activations and tested zero-shot on the T/F evaluation tokens of HaluEval (top) and PopQA (bottom).}
    \label{fig:app_layerwise_tf}
\end{figure}

\begin{figure}[H]
    \centering
    \includegraphics[width=\textwidth]{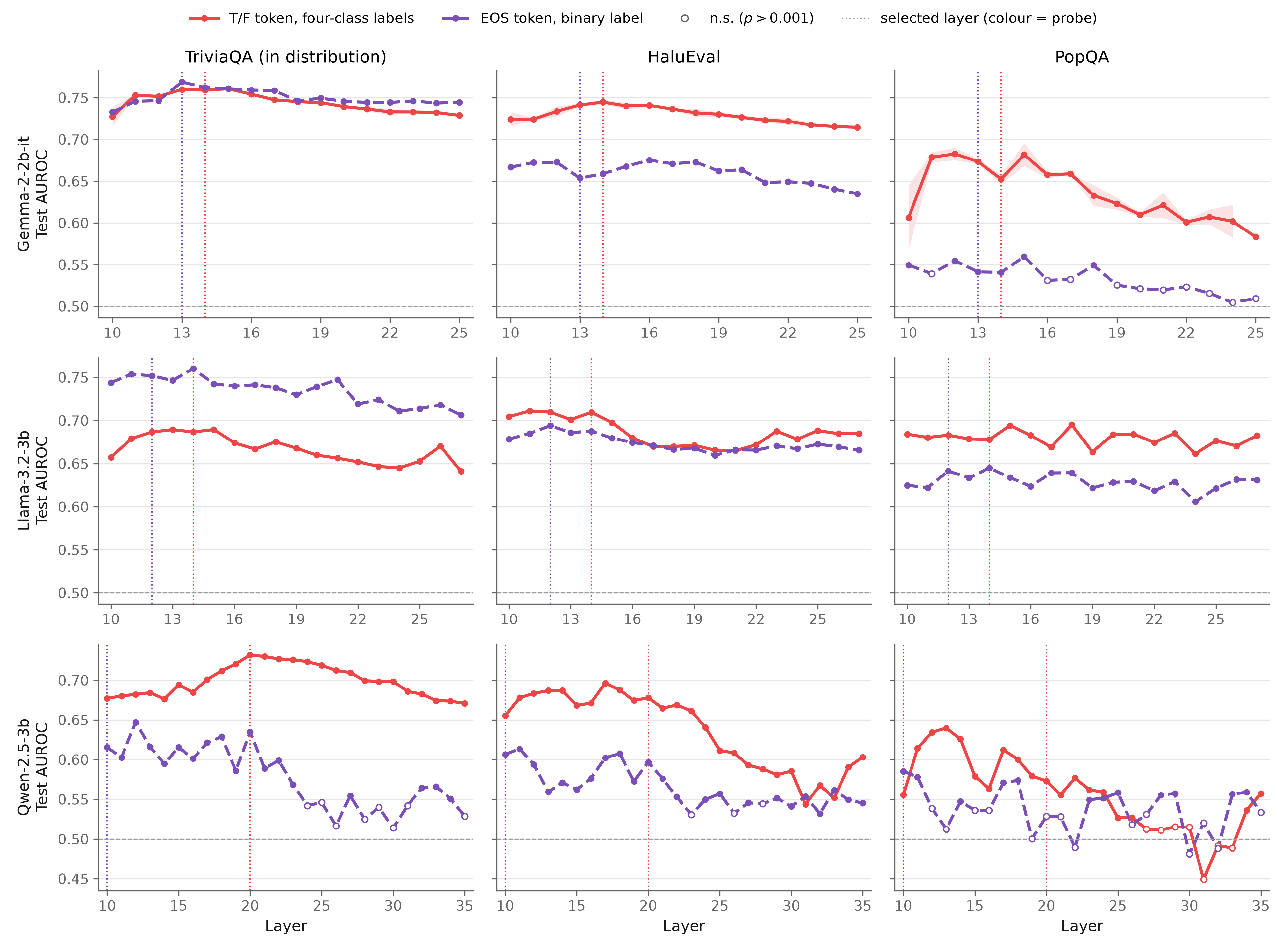}
    \caption{Layer-wise test AUROC of four-class T/F MLP probes (solid; trained and tested on T/F evaluation tokens) and binary EOS MLP probes (dashed; trained and tested on EOS tokens with a correct/incorrect label), all trained on TriviaQA. Rows: models; columns: test set.}
    \label{fig:confabulation}
\end{figure}

This appendix collects layer-wise test AUROC plots for cross-dataset transfer of T/F probes (Figure~\ref{fig:app_layerwise_tf}) and the four-class T/F versus binary EOS comparison across all three models (Figure~\ref{fig:confabulation}).

\clearpage
\section{Validation AUROC by Layer}\label{app:val_layerwise}

\begin{figure}[H]
    \centering
    \includegraphics[width=\textwidth]{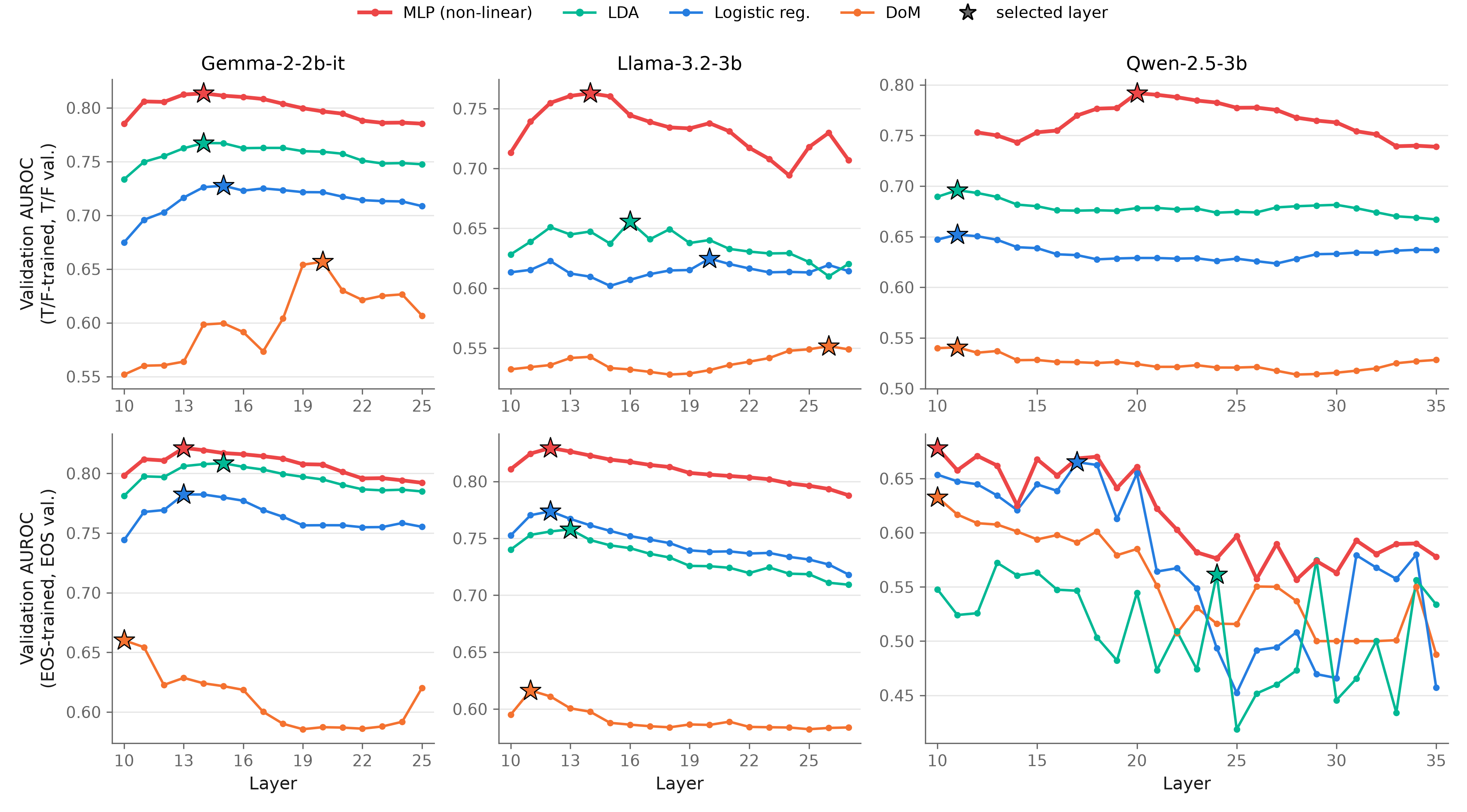}
    \caption{Layer-wise validation AUROC (TriviaQA) used for layer selection. Top: T/F-trained probes on the T/F validation set. Bottom: EOS-trained probes on the EOS validation set. Stars mark each probe's selected layer (highest validation AUROC among layers $\geq 10$ significant on TriviaQA test, $p \leq 0.001$). Qwen-2.5-3b MLP validation scores are available from layer 12. EOS-trained LDA for Llama-3.2-3b and Qwen-2.5-3b uses a 20K subsample.}
    \label{fig:val_layerwise}
\end{figure}

Figure~\ref{fig:val_layerwise} shows the validation AUROC of every probe at every layer $\geq 10$, with the selected layer marked.

\clearpage
\section{T/F-to-EOS Transfer Across Models}\label{app:eos_transfer_other}

\begin{figure}[H]
    \centering
    \includegraphics[width=\textwidth]{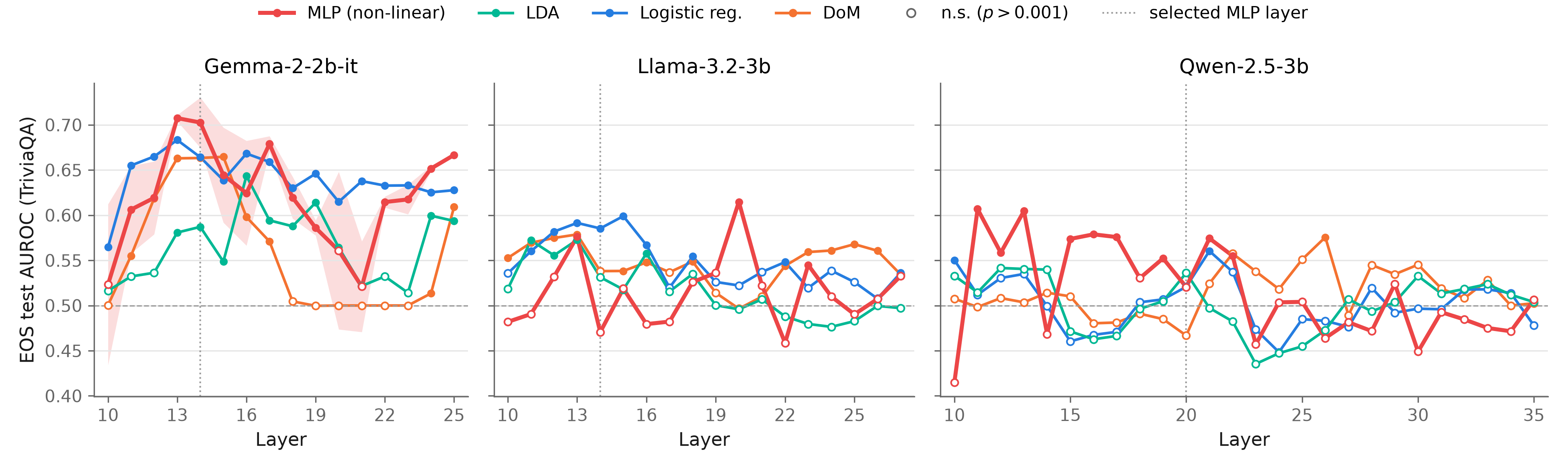}
    \caption{Layer-wise test AUROC of probes trained on TriviaQA T/F activations and tested without retraining on TriviaQA EOS activations (no evaluation token).}
    \label{fig:eos_transfer}
\end{figure}

Table~\ref{tab:eos_transfer_other} and Figure~\ref{fig:eos_transfer} present the full T/F-to-EOS cross-position transfer results across all three models, where probes trained on TriviaQA T/F activations are evaluated without retraining on TriviaQA EOS activations (without any appended evaluation token).

\begin{table}[H]
\centering
\small
\begin{tabular}{ll c c c}
\toprule
\textbf{Model} & \textbf{Probe} & \textbf{Layer} & \textbf{T/F Val AUROC} & \textbf{EOS Test AUROC} \\
\midrule
\multirow{4}{*}{\texttt{Gemma-2-2b-it}}
 & Non-linear (MLP) & 14 & 0.8135 & \textbf{0.7026} \\
 & Linear (LogReg)  & 15 & 0.7276 & 0.6387 \\
 & Linear (LDA)     & 14 & 0.7672 & 0.5872 \\
 & Linear (DoM)     & 20 & 0.6569 & n.s. \\
\midrule
\multirow{4}{*}{\texttt{Llama-3.2-3b}}
 & Non-linear (MLP) & 14 & 0.7628 & n.s. \\
 & Linear (LogReg)  & 20 & 0.6248 & n.s. \\
 & Linear (LDA)     & 16 & 0.6557 & 0.5578 \\
 & Linear (DoM)     & 26 & 0.5517 & 0.5606 \\
\midrule
\multirow{4}{*}{\texttt{Qwen-2.5-3b}}
 & Non-linear (MLP) & 20 & 0.7916 & n.s. \\
 & Linear (LogReg)  & 11 & 0.6520 & n.s. \\
 & Linear (LDA)     & 11 & 0.6959 & n.s. \\
 & Linear (DoM)     & 11 & 0.5406 & n.s. \\
\bottomrule
\end{tabular}
\caption{T/F-trained probes evaluated on the EOS token (TriviaQA test) at the layer selected on T/F validation. n.s.: not significant ($p > 0.001$). Transfer is strong on Gemma-2-2b-it (MLP 0.7026 AUROC), weak on Llama-3.2-3b (only LDA and DoM significant, near 0.56), and not significant for any probe on Qwen-2.5-3b.}
\label{tab:eos_transfer_other}
\end{table}

\section{Probe Training and Evaluation Details}\label{app:training}

\subsection{Non-Linear Probe Architecture}

The non-linear probe is a multi-layer perceptron (MLP) with three hidden layers,
\[
\phi \xrightarrow{W_1} \text{LeakyReLU} \xrightarrow{W_2} \text{LeakyReLU} \xrightarrow{W_3} \text{LeakyReLU} \xrightarrow{W_4} \text{logit},
\]
with hidden widths 256, 64 and 32. We chose this architecture after manually comparing several layer widths and activation functions, and use it for every model, layer and position (and, with input size $k$, for the SVD probes of Appendix~\ref{app:svd_details}).

\subsection{Training and Hyperparameter Tuning}

Probes are trained with binary cross-entropy, AdamW (weight decay $10^{-4}$), a linear warmup over the first 10\% of steps followed by cosine decay, and early stopping on validation AUROC (at most 50 epochs), keeping the weights of the best epoch. For each (model, layer, activation position), the learning rate and batch size are tuned with Optuna~\citep{akiba2019optuna} over $\{10^{-5}, 10^{-4}, 10^{-3}, 10^{-2}\} \times \{64, 128, 256, 512\}$, and the probe of the best trial is kept. For Gemma-2-2b-it T/F probes, we repeat this for three seeds (42, 43, 44) and report the mean AUROC and permutation $p$-value over seeds. All other MLP probes (Gemma-2-2b-it EOS, and all Llama-3.2-3b and Qwen-2.5-3b probes) use a single seed. Linear probes are deterministic given the training data. The validation set is used only for early stopping, hyperparameter selection and layer selection (Section~\ref{ssec:probing}); it is never used to compute reported test scores.

\subsection{Linear Probe Details}

Logistic regression is trained on \texttt{StandardScaler}-normalised activations with an L2 penalty ($C=1.0$), using L-BFGS for up to 2000 iterations. Difference-of-means (DoM) uses raw activations with no preprocessing: $w_{\text{DoM}} = \hat{\mu}_1 - \hat{\mu}_0$. AUROC is computed from the decision function (logistic regression, LDA) or the sigmoid of the projection onto $w_{\text{DoM}}$ (DoM). Because AUROC depends only on the ranking of scores, the sigmoid does not affect it.

\subsection{LDA: Training Set Size and Shrinkage}\label{app:lda_shrinkage}

\citet{lee2026meanshift} argue that apparent gains of non-linear probes over linear ones can stem from poor covariance estimation, and that shrinkage LDA recovers much of the gap. We therefore examined how the LDA baseline depends on the amount of training data and on covariance shrinkage.

\paragraph{LDA.} LDA models the activations of each class $k \in \{0, 1\}$ as Gaussian with mean $\mu_k$ and a covariance $\Sigma$ shared by both classes, $\phi \mid z{=}k \sim \mathcal{N}(\mu_k, \Sigma)$, with class prior $\pi_k$. The quadratic terms of the two Gaussians then cancel, and the log-odds are linear in $\phi$:
\begin{align*}
\log \frac{P(z{=}1 \mid \phi)}{P(z{=}0 \mid \phi)} &= w^\top \phi + b, \qquad w = \Sigma^{-1}(\mu_1 - \mu_0), \\
b &= -\tfrac{1}{2}\left(\mu_1^\top \Sigma^{-1} \mu_1 - \mu_0^\top \Sigma^{-1} \mu_0\right) + \log \frac{\pi_1}{\pi_0}.
\end{align*}
The probe score is $\sigma(w^\top \phi + b)$. Compared with the difference of means, $w_{\text{DoM}} = \mu_1 - \mu_0$, LDA reweights the mean difference by $\Sigma^{-1}$, down-weighting directions in which the activations vary strongly for reasons unrelated to the label. In practice $\mu_k$ and $\Sigma$ are replaced by their sample estimates $\hat{\mu}_k$ and $\hat{\Sigma}$ from $n$ training examples in $d$ dimensions ($d = 2304$, $3072$, $2048$ for Gemma-2-2b-it, Llama-3.2-3b and Qwen-2.5-3b).

\paragraph{Why the training set size matters.} $\hat{\Sigma}$ has $d(d+1)/2$ free parameters (2.4--4.7M here) and is inverted to obtain $w$, so errors in its estimate propagate into the probe direction. Errors are largest in the smallest eigenvalues: even when the true covariance is the identity, the eigenvalues of the sample covariance spread over the interval $\left[(1-\sqrt{d/n})^2,\ (1+\sqrt{d/n})^2\right]$ as $n, d \to \infty$~\citep{marchenko1967distribution}, and $\hat{\Sigma}^{-1}$ amplifies the underestimated directions most. With a 20{,}000-example subsample, $n/d \approx 6.5$ (Llama-3.2-3b) to $9.8$ (Qwen-2.5-3b), and the interval is roughly $[0.37, 1.94]$ for Llama-3.2-3b, a five-fold spread in a matrix that should have all eigenvalues equal to 1. With the full training sets, $n/d \approx 280$ (Gemma-2-2b-it), $400$ (Llama-3.2-3b) and $600$ (Qwen-2.5-3b), and the interval shrinks to within about $\pm 10\%$ of 1 (e.g.\ $[0.90, 1.10]$ for Llama-3.2-3b).

\paragraph{Shrinkage.} Shrinkage stabilises $\hat{\Sigma}$ by pulling it towards a well-conditioned target. Following scikit-learn's implementation, we standardise each class with its own feature-wise standard deviations $D_k = \operatorname{diag}(\hat{\sigma}_k)$, compute the class correlation matrix $\hat{R}_k = \frac{1}{n_k}\sum_{i:\,z_i=k} u_i u_i^\top$ with $u_i = D_k^{-1}(\phi_i - \hat{\mu}_k)$, shrink it towards the identity, and pool the rescaled class covariances:
\[
\hat{\Sigma}_\alpha = \sum_{k \in \{0,1\}} \hat{\pi}_k \, D_k \left[(1-\alpha)\,\hat{R}_k + \alpha\, \tfrac{\operatorname{tr}(\hat{R}_k)}{d}\, I\right] D_k, \qquad \alpha \in [0, 1].
\]
Since $\hat{R}_k$ has a unit diagonal, $\operatorname{tr}(\hat{R}_k)/d = 1$. At $\alpha = 0$ this is the unshrunk LDA. At $\alpha = 1$, $\hat{\Sigma}_1 = \sum_k \hat{\pi}_k D_k^2$ is diagonal, and the probe becomes a \emph{diagonal} LDA, $w_j = (\hat{\mu}_{1j} - \hat{\mu}_{0j})/\hat{\sigma}_j^2$, which ignores all correlations between features. The Ledoit--Wolf estimator~\citep{ledoit2004well} chooses $\alpha$ analytically to minimise the expected squared error of the shrunk estimate: for each class, with $\hat{m} = \operatorname{tr}(\hat{R})/d$,
\[
\alpha_{\text{LW}} = \frac{\min(\beta, \delta)}{\delta}, \qquad
\delta = \frac{1}{d}\left\lVert \hat{R} - \hat{m} I \right\rVert_F^2, \qquad
\beta = \frac{1}{d\,n}\left(\frac{1}{n}\sum_{i} \lVert u_i \rVert^4 - \lVert \hat{R} \rVert_F^2\right),
\]
where $\delta$ measures how far $\hat{R}$ is from the target and $\beta$ estimates the sampling variance of $\hat{R}$; shrinkage is strong only when the sampling noise is large relative to the structure being shrunk. This is identical to scikit-learn's \texttt{LinearDiscriminantAnalysis(solver="lsqr", shrinkage="auto")}. Because the full training sets do not fit in memory, we accumulate $\hat{\mu}_k$, $\hat{\sigma}_k$, $\hat{R}_k$ and $\sum_i \lVert u_i \rVert^4$ in two streaming passes over the data.

\paragraph{Variants.} We compare four variants:
\begin{itemize}
    \item \textbf{Subsample}: LDA ($\alpha = 0$) fitted on a random 20{,}000-example subsample (our initial baseline for Llama-3.2-3b and Qwen-2.5-3b).
    \item \textbf{Full data}: LDA ($\alpha = 0$) fitted on the full balanced training set (643K examples for Gemma-2-2b-it, 1.22M for Llama-3.2-3b and Qwen-2.5-3b, the same training sets used for all other probes). This is the LDA reported in the main text.
    \item \textbf{Ledoit--Wolf}: full data, $\alpha = \alpha_{\text{LW}}$.
    \item \textbf{Validation-tuned shrinkage}: full data, $\alpha$ chosen by validation AUROC from $\{0, 10^{-4}, 3\cdot10^{-4}, 10^{-3}, 3\cdot10^{-3}, 0.01, 0.02, 0.05, 0.1, 0.2, 0.3, 0.5, 0.7, 0.9, 1\}$.
\end{itemize}
\paragraph{Training set size matters; shrinkage does not.} Table~\ref{tab:lda_shrinkage} shows that fitting LDA on the full data raises TriviaQA test AUROC by 0.054 for Llama-3.2-3b and 0.033 for Qwen-2.5-3b, closing 41\% and 25\% of the gap to the MLP at the selected layers. Once the full data are used, shrinkage has almost no effect. The Ledoit--Wolf coefficient is $\alpha_{\text{LW}} \approx 2$--$8 \times 10^{-5}$ at every layer of every model, i.e.\ effectively zero, as expected when $n/d$ is in the hundreds. Tuning $\alpha$ on validation selects small values for Gemma-2-2b-it and Qwen-2.5-3b ($\leq 0.01$) and moderate ones for Llama-3.2-3b (0.01--0.3); at the selected layers it changes test AUROC by at most 0.010 (Llama-3.2-3b) and by less than 0.001 for the other two models. Beyond these values, validation AUROC falls steadily as $\alpha$ grows, down to 0.52--0.65 at $\alpha = 1$ (the diagonal limit), compared with 0.61--0.77 at $\alpha = 0$. The correlations between features captured by $\hat{\Sigma}$ are therefore essential to the linear probe, but once they are estimated from the full data, the remaining gap to the MLP is not explained by covariance estimation error.

\begin{table}[H]
\centering
\footnotesize
\setlength{\tabcolsep}{4.5pt}
\begin{tabular}{l l c c ccc}
\toprule
\textbf{Model} & \textbf{Probe} & \textbf{Layer} & \textbf{Val AUROC} & \multicolumn{3}{c}{\textbf{Test AUROC}} \\
\cmidrule(lr){5-7}
 & & & TriviaQA & TriviaQA & HaluEval & PopQA \\
\midrule
\multirow{3}{*}{\texttt{Gemma-2-2b-it}}
 & LDA, full data           & 14 & 0.7672 & 0.7113 & 0.7032 & 0.5576 \\
 & LDA, val.-tuned $\alpha$ & 14 & 0.7673 & 0.7116 & 0.7032 & 0.5579 \\
 & MLP                      & 14 & 0.8135 & 0.7591 & 0.7447 & 0.6525 \\
\midrule
\multirow{4}{*}{\texttt{Llama-3.2-3b}}
 & LDA, 20K subsample       & 14 & 0.5928 & 0.5568 & 0.5780 & 0.5802 \\
 & LDA, full data           & 16 & 0.6557 & 0.6104 & 0.6304 & 0.6614 \\
 & LDA, val.-tuned $\alpha$ & 16 & 0.6601 & 0.6171 & 0.6404 & 0.6698 \\
 & MLP                      & 14 & 0.7628 & 0.6867 & 0.7095 & 0.6778 \\
\midrule
\multirow{4}{*}{\texttt{Qwen-2.5-3b}}
 & LDA, 20K subsample       & 11 & 0.6491 & 0.5990 & 0.6213 & 0.5890 \\
 & LDA, full data           & 11 & 0.6959 & 0.6318 & 0.6431 & 0.5954 \\
 & LDA, val.-tuned $\alpha$ & 11 & 0.6959 & 0.6318 & 0.6431 & 0.5954 \\
 & MLP                      & 20 & 0.7916 & 0.7316 & 0.6779 & 0.5731 \\
\bottomrule
\end{tabular}
\caption{LDA variants on T/F activations, each at its layer selected on TriviaQA T/F validation (Section~\ref{ssec:probing}). For Gemma-2-2b-it the original LDA was already fitted on the full training set. The full-data variants were fitted at every layer $\geq 10$ for Llama-3.2-3b and Qwen-2.5-3b, and the shrinkage variants at layers 12--17 for Gemma-2-2b-it. Ledoit--Wolf results match the full-data LDA to within 0.001 and are omitted. All entries significant ($p \leq 0.001$).}
\label{tab:lda_shrinkage}
\end{table}

\section{Dataset Details}\label{app:data}

\subsection{Dataset Statistics}\label{app:data_stats}

\begin{table}[H]
\centering
\footnotesize
\setlength{\tabcolsep}{3.5pt}
\begin{tabular}{l r rr rr rr rr}
\toprule
 & & \multicolumn{2}{c}{\textbf{Validation}} & \multicolumn{2}{c}{\textbf{TriviaQA test}} & \multicolumn{2}{c}{\textbf{HaluEval test}} & \multicolumn{2}{c}{\textbf{PopQA test}} \\
\cmidrule(lr){3-4}\cmidrule(lr){5-6}\cmidrule(lr){7-8}\cmidrule(lr){9-10}
\textbf{Model} & \textbf{Train} & T/F & EOS & T/F & EOS & T/F & EOS & T/F & EOS \\
\midrule
\texttt{Gemma-2-2b-it} & 643,188   & 126,668 & 63,334 & 5,248 & 2,624 & 1,712 & 2,572 & 5,296 & 2,343 \\
\texttt{Llama-3.2-3b}  & 1,221,092 & 121,716 & 60,876 & 6,248 & 3,124 & 5,834 & 2,918 & 1,420 & 2,883 \\
\texttt{Qwen-2.5-3b}   & 1,224,744 & 120,468 & 60,168 & 5,636 & 1,184 & 5,390 & 2,695 & 5,502 & 2,751 \\
\bottomrule
\end{tabular}
\caption{Activation--label pair counts per layer across splits.}
\label{tab:data_stats}
\end{table}

Training, validation and test examples come from disjoint TriviaQA splits of 20{,}000, 2{,}000 and 2{,}000 queries, with 32 sampled generations per query. In the training and validation sets, we drop exact duplicate generations for the same query. In the test sets (TriviaQA test, HaluEval and PopQA), we keep the first correct and the first incorrect generation of each query (only one, if all 32 samples are correct or all are incorrect), so no query is over-represented. Each generation yields one EOS example and two T/F examples (one per suffix), so a T/F test set built from the same generations is twice the size of the corresponding EOS set; this holds for TriviaQA on Gemma-2-2b-it and Llama-3.2-3b. The remaining T/F and EOS test sets differ in size (for example, the Qwen-2.5-3b TriviaQA EOS set contains 1,184 examples, obtained using a subset of the queries in T/F). The training and validation sets are balanced over the binary target $z$. The T/F test sets are balanced by construction, since each generation contributes one example with $z{=}1$ and one with $z{=}0$. The EOS test sets keep their natural ratio of correct to incorrect generations; this does not affect AUROC. HaluEval and PopQA are used only as test sets, with no retraining and no layer re-selection; probe layers are always chosen on TriviaQA validation.

\subsection{Label Quality: Clarification Requests}\label{app:grounding}

Correctness labels come from fuzzy matching between each generation and the reference answer (Section~\ref{ssec:representation}). Some generations do not answer the question but ask for clarification instead (for example, which of several people with the same name is meant). Fuzzy matching labels these as incorrect, although they are not factual errors. Table~\ref{tab:clarification} reports how often they occur; they were identified by an automated pass over all generations with an LLM (Claude Opus 5.5).

\begin{table}[H]
\centering
\small
\begin{tabular}{ll rrr}
\toprule
\textbf{Model} & \textbf{Dataset} & \textbf{Clarification gens} & \textbf{\% of gens} & \textbf{Questions affected} \\
\midrule
\multirow{5}{*}{\texttt{Gemma-2-2b-it}}
 & TriviaQA train$^\dagger$ & 1{,}596 / 412{,}962 & 0.39 & 215 \\
 & TriviaQA val   & 77 / 64{,}000      & 0.12 & 21 \\
 & TriviaQA test  & 194 / 64{,}000     & 0.30 & 27 \\
 & PopQA          & 2{,}211 / 55{,}072 & 4.0  & 224 \\
 & HaluEval       & 1{,}636 / 64{,}000 & 2.6  & 170 \\
\midrule
\multirow{5}{*}{\texttt{Llama-3.2-3b}}
 & TriviaQA train & 5{,}350 / 640{,}000 & 0.84 & 1{,}786 \\
 & TriviaQA val   & 562 / 64{,}000      & 0.88 & 175 \\
 & TriviaQA test  & 612 / 64{,}000      & 0.96 & 176 \\
 & PopQA          & 13{,}709 / 63{,}200 & \textbf{21.7} & 1{,}291 \\
 & HaluEval       & 3{,}931 / 64{,}000  & 6.1  & 693 \\
\midrule
\multirow{5}{*}{\texttt{Qwen-2.5-3b}}
 & TriviaQA train & 7{,}581 / 640{,}000 & 1.2  & 1{,}972 \\
 & TriviaQA val   & 853 / 64{,}000      & 1.33 & 208 \\
 & TriviaQA test  & 863 / 64{,}000      & 1.35 & 190 \\
 & PopQA          & 9{,}828 / 63{,}200  & \textbf{15.6} & 1{,}040 \\
 & HaluEval       & 3{,}739 / 64{,}000  & 5.8  & 598 \\
\bottomrule
\end{tabular}
\caption{Generations that request clarification instead of answering, as a share of all sampled generations (32 per question), and the number of questions with at least one such generation. $^\dagger$Counted over unique generations.}
\label{tab:clarification}
\end{table}

Clarification requests are rare in TriviaQA (at most 1.35\% of generations) but more common in HaluEval (2.6--6.1\%) and especially in PopQA (4.0\% for Gemma-2-2b-it, 15.6\% for Qwen-2.5-3b and 21.7\% for Llama-3.2-3b). PopQA asks about long-tail entities, often with ambiguous names, which plausibly prompts the models to ask which entity is meant. Because the probes are trained on TriviaQA, where such responses are almost absent, part of the PopQA incorrect class is a type of response the probes have rarely seen labelled as incorrect. This may contribute to the lower transfer to PopQA (Sections~\ref{ssec:cross-dataset} and~\ref{sec:subspace}).

\section{SVD Probe Details}\label{app:svd_details}

For each layer $\ell$, we compute the SVD $\Phi_\ell = U S V^\top$ of the training activation matrix $\Phi_\ell \in \mathbb{R}^{N \times d}$ (no mean-centring) and project an activation $\phi$ onto the top-$k$ right singular vectors, $z = V_{:, 1:k}^\top \phi \in \mathbb{R}^k$. For $k \leq 256$, the bases are nested: we compute the top 256 right singular vectors exactly (as the leading eigenvectors of $\Phi_\ell^\top \Phi_\ell$), and the $k$-dimensional basis is their first $k$ columns, so smaller $k$ always keeps the directions with the largest singular values. For $k = 576$, the basis is computed separately with a randomized truncated SVD. The basis is computed only on training data and applied unchanged to validation and test activations. Every SVD probe is an MLP with input size $k$ ($k \to 256 \to 64 \to 32 \to 1$, LeakyReLU), trained with binary cross-entropy, AdamW (weight decay $10^{-4}$), a linear warmup over the first 10\% of steps followed by cosine decay, and early stopping on validation AUROC (up to 50 epochs, keeping the best epoch's weights), with one seed. Probes with $k \leq 256$ use a fixed recipe (learning rate $10^{-4}$, batch size 512, patience 3), taken from the most frequent best setting of a preliminary search at $k = 256$. For $k = 576$, the learning rate and batch size are tuned per layer with Optuna over $\{10^{-5}, 10^{-4}, 10^{-3}, 10^{-2}\} \times \{64, 128, 256, 512\}$. Evaluation uses the same unique-only test sets and permutation test as the full-dimensional probes.

\section{Layer-wise SVD Probe Results}\label{app:svd_layerwise}

\begin{figure}[H]
    \centering
    \includegraphics[width=\textwidth]{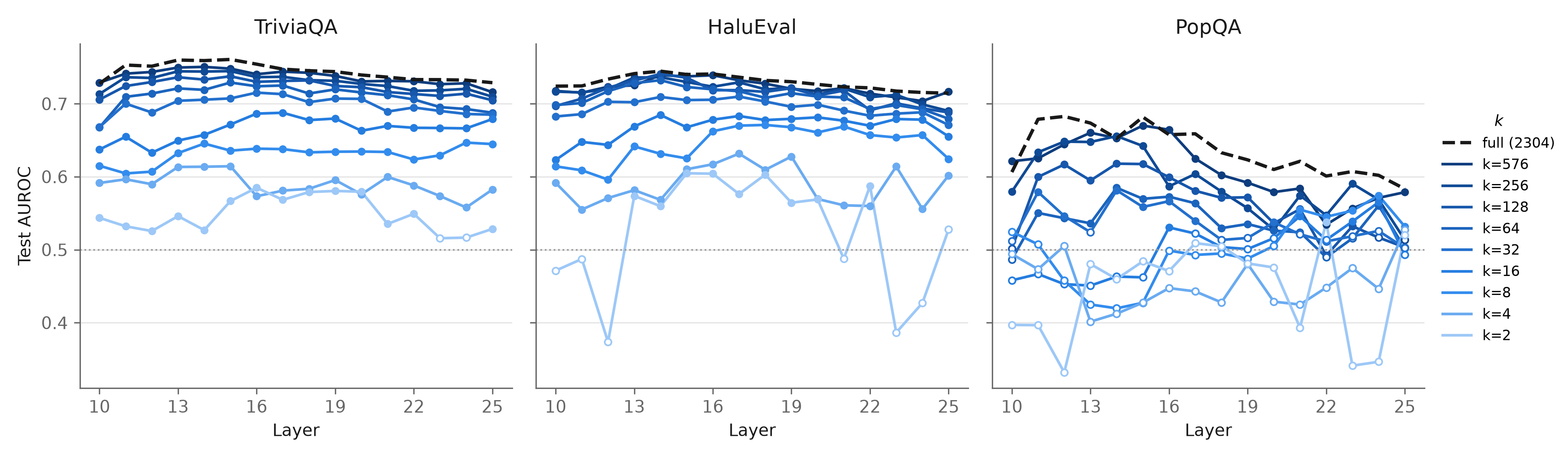}
    \caption{Layer-wise test AUROC of SVD-projected non-linear probes (Gemma-2-2b-it, trained on TriviaQA T/F) on TriviaQA (left), HaluEval (middle) and PopQA (right). Darker blue: larger $k$; dashed black: full-dimensional probe (mean over three seeds). Filled markers: significant ($p \leq 0.001$); hollow: not significant. Dotted line: chance.}
    \label{fig:svd_layerwise}
\end{figure}

Figure~\ref{fig:svd_layerwise} shows the test AUROC of the SVD-projected probes at every layer from 10 to 25, for all $k$, together with the full-dimensional probe. On TriviaQA and HaluEval, the curves are ordered by $k$ at almost every layer, and the probes with $k \geq 128$ stay close to the full probe throughout. On PopQA, the probes with $k \leq 16$ are near or below chance at most layers: the number of layers (out of 16) that are not significant is 0, 2, 4, 4 and 9 for $k = 576, 256, 128, 64, 32$, and 12, 11, 16 and 15 for $k = 16, 8, 4, 2$. AUROCs below 0.5 occur only on PopQA (at 7, 8, 14 and 12 layers for $k = 16, 8, 4, 2$) and on HaluEval at $k = 2$ (6 layers). Because the permutation test is one-sided, these below-chance values are reported as not significant, although they indicate a systematic inversion rather than an absence of signal.

\section{SHAP Feature Attribution}\label{app:shap}

We apply SHAP with gradient-based attribution (\texttt{GradientExplainer})~\citep{lundberg2017unified} to the non-linear probes trained in the $k = 576$ SVD subspace. For each sample $i$ and SVD direction $j$, the SHAP value $\phi_j^{(i)}$ measures how much direction $j$ contributed to the probe's output for that sample.

\begin{figure}[H]
    \centering
    \includegraphics[width=0.55\textwidth]{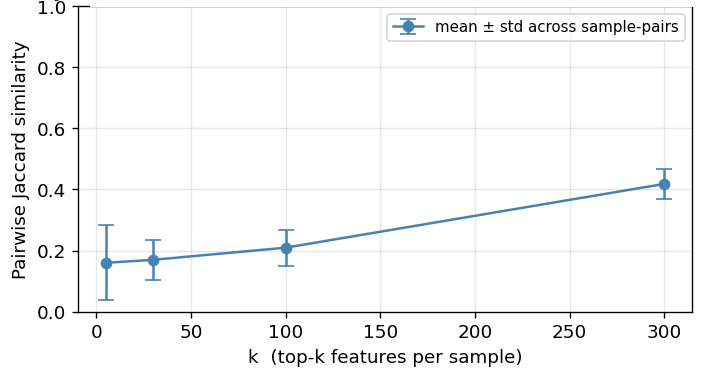}
    \caption{Mean pairwise Jaccard similarity between per-sample top-$m$ SHAP direction sets ($\pm$ std over sample pairs), for the $k = 576$ SVD probe on Gemma-2-2b-it, layer 14 ($N_{\text{sub}} = 200$). Chance levels for random $m$-subsets of 576 directions are 0.004, 0.03, 0.10 and 0.35 at $m = 5, 30, 100, 300$. In the plots, $k$ denotes the number of top directions per sample ($m$ in the text).}
    \label{fig:shap_jaccard_vs_k}
\end{figure}

\paragraph{Consistency across samples.} We test whether different inputs rely on the same SVD directions. For $N_{\text{sub}} = 200$ test samples, we take each sample's top-$m$ directions by $|\phi_j^{(i)}|$ and compute the Jaccard similarity between every pair of samples (Figure~\ref{fig:shap_jaccard_vs_k}). Two random $m$-subsets of $D$ directions have an expected Jaccard of roughly $(m^2/D)/(2m - m^2/D)$, which we use as the chance level. For small $m$ the overlap is far above chance: 0.16 at $m = 5$, against 0.004 for random sets in $D = 576$ dimensions, and 0.17 at $m = 30$, against 0.03. Different inputs therefore share a small core of dominant directions. For larger $m$, the overlap approaches chance (0.42 vs.\ 0.35 at $m = 300$), so agreement beyond this core is weak.

\begin{figure}[H]
    \centering
    \begin{minipage}[b]{0.48\textwidth}
        \centering
        \includegraphics[width=\textwidth]{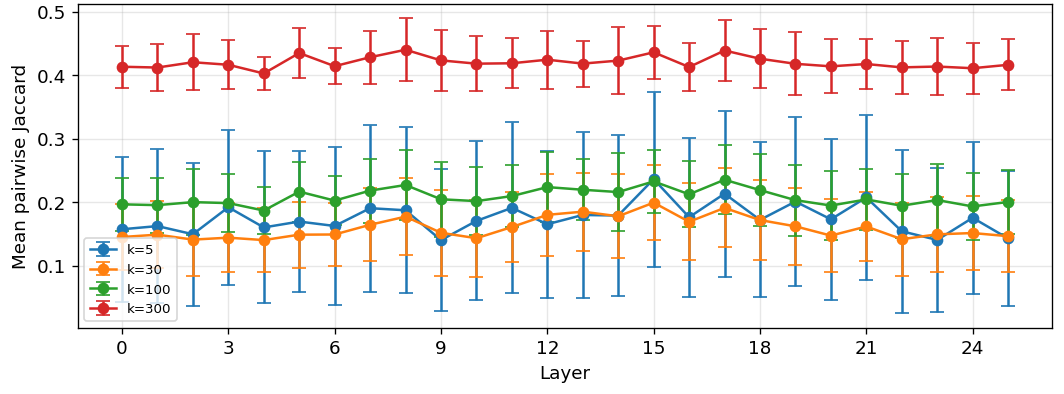}
        \subcaption*{SVD probe: Jaccard by layer and $m$}
    \end{minipage}
    \hfill
    \begin{minipage}[b]{0.48\textwidth}
        \centering
        \includegraphics[width=\textwidth]{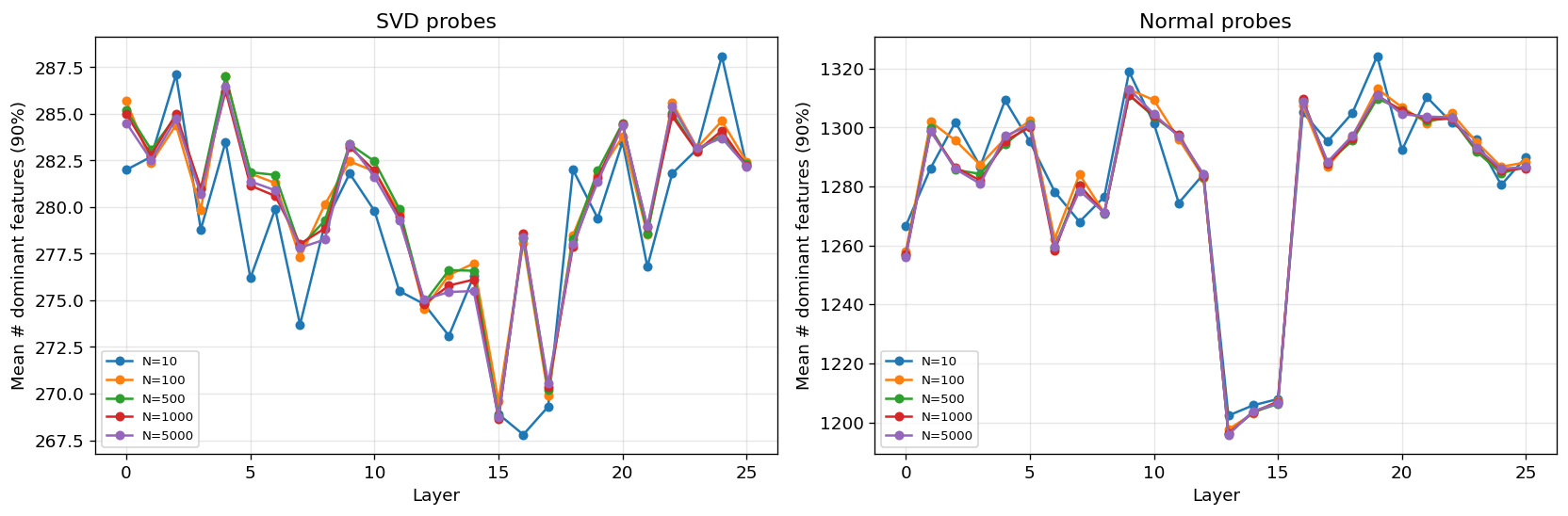}
        \subcaption*{SVD vs.\ full-space: dominant feature count (90\%)}
    \end{minipage}
    \caption{\textbf{Left:} Mean pairwise Jaccard similarity of per-sample top-$m$ SHAP sets across layers, for $k = 576$ SVD probes (Gemma-2-2b-it, T/F). \textbf{Right:} Number of features needed to cover 90\% of $|$SHAP$|$ mass for SVD probes ($\sim$270--290 of 576) and full-space probes ($\sim$1200--1330 of 2304). In the plots, $k$ denotes the number of top directions per sample ($m$ in the text).}
    \label{fig:shap_jaccard_by_layer}
\end{figure}

\paragraph{Across layers, and SVD vs.\ full-space probes.} Figure~\ref{fig:shap_jaccard_by_layer} (left) shows that the pairwise Jaccard pattern is similar at all layers. The right panel compares the number of features needed to cover 90\% of the $|$SHAP$|$ mass: about 270--290 of 576 directions for SVD probes and about 1200--1330 of 2304 dimensions for full-space probes. As a fraction of the input dimension, these are similar (about half the features in both cases), so the SHAP mass is not more concentrated in the SVD basis than in the original basis; the smaller absolute count reflects the smaller input dimension. Jaccard values are likewise not directly comparable between the two settings, because the chance level depends on the dimension (0.35 for $m = 300$ in 576 dimensions vs.\ 0.07 in 2304).

\section{Code and Data Availability}\label{app:code}

Code for generation, activation extraction, probe training, evaluation and permutation testing is available at \url{https://github.com/AI-Club-IIT-Madras/Generalizable-Probes-for-Detecting-Factual-Correctness-in-LLMs}. TriviaQA, HaluEval and PopQA are publicly available. Model-generated completions, and all evaluation results are available at \url{https://drive.google.com/drive/folders/1MYy4VNIONU0sUAcNAtM1c9rHKxohs0Fh}.

\end{document}